\documentclass{article}

\usepackage[dblblindworkshop,final]{neurips_2026}

\usepackage[utf8]{inputenc} %
\usepackage[T1]{fontenc}    %
\usepackage{hyperref}       %
\usepackage{url}            %
\usepackage{booktabs}       %
\usepackage{amsfonts}       %
\usepackage{nicefrac}       %
\usepackage{microtype}      %
\usepackage{xcolor}         %

\usepackage{graphicx}
\usepackage{tabularx}
\usepackage{subfigure}
\usepackage{booktabs}
\hypersetup{hidelinks}

\usepackage{amsmath}
\usepackage{amssymb}
\usepackage{mathtools}
\usepackage{amsthm}
\usepackage[capitalize,noabbrev]{cleveref}

\usepackage{multirow}
\usepackage{tabularx}
\usepackage{url}

\theoremstyle{plain}

\theoremstyle{definition}

\theoremstyle{remark}

\usepackage{xcolor}
\usepackage[most]{tcolorbox}

\newtcolorbox{researchquestionbox}{
  enhanced,
  breakable,
  frame hidden,
  boxrule=0pt,
  colback=gray!4,
  borderline west={2pt}{0pt}{violet!70!black},
  sharp corners,
  left=7pt,
  right=5pt,
  top=5pt,
  bottom=5pt,
  before skip=0.7em,
  after skip=0.8em,
  fontupper=\itshape
}

\newtcolorbox{takeawaybox}{
  enhanced,
  breakable,
  colback=teal!5!white,
  colframe=white,
  coltitle=white,
  fonttitle=\bfseries,
  attach boxed title to top left={xshift=6pt,yshift=-2pt},
  borderline west={2pt}{0pt}{orange!85!black},
  boxrule=0.5pt,
  arc=2pt,
  left=6pt,
  right=6pt,
  top=8pt,
  bottom=6pt,
  before skip=0.9em,
  after skip=1.0em
}

\usepackage{pgfplotstable}
\usepackage{booktabs}
\usepackage{array}

\pgfplotsset{compat=1.18}

\pgfplotstableset{
  csvtable/.style={
    col sep=comma,
    string type,
    every head row/.style={
      before row=\toprule,
      after row=\midrule
    },
    every last row/.style={
      after row=\bottomrule
    }
  }
}

\usepackage[table]{xcolor}
\usepackage{makecell}

\newcommand{\metriccol}[1]{\cellcolor{teal!18}#1}
\newcommand{\bestmetric}[1]{\cellcolor{teal!30}\textbf{#1}}

\title{How Do Video Foundation Models Encode Intuitive Physics? Probing Across Pretraining Paradigms}

\workshoptitle{World Models in Physical AI}

\author{%
  Samuele Punzo$^{*}$ \\
  University of Amsterdam
  \And
  Niccolò Caselli$^{*}$ \\
  University of Amsterdam
  \And
  Ippokratis Pantelidis$^{*}$ \\
  University of Amsterdam
  \AND
  Francesco Massafra$^{*}$ \\
  University of Amsterdam
  \And
  Salvatore Lo Sardo$^{*}$ \\
  University of Amsterdam
  \And
  Mohammadreza Salehi$^{\dagger}$ \\
  University of Amsterdam
}
\begin{document}

\maketitle

\begingroup
\renewcommand{\thefootnote}{}
\footnotetext{%
  $^{*}$Equal contribution.
  $^{\dagger}$Supervisor.
}
\endgroup

\begin{abstract}
\textcolor{black}{\textit{Do video foundation models encode a usable sense of how the physical world works, and does intuitive physics emerge naturally from large-scale video pretraining?} In this work,} we study whether pretrained video foundation models encode intuitive-physics information in their frozen representations, and how its accessibility varies across \textcolor{black}{pretraining paradigms}, layers, backbone variants and probe types. Using frozen-feature probing on IntPhys2 and Minimal Video Pairs (MVP), we compare predictive joint-embedding models (V-JEPA), masked reconstruction models (VideoMAE), and diffusion-based video generators (LTX-Video). \textcolor{black}{We find that across linear, MLP, and temporal attentive probes, physics-relevant information is generally weakest in early layers and most accessible at intermediate to late depth. On MVP, temporal attentive probes substantially outperform linear readouts, with V-JEPA achieving 95.0\% pair consistency. On IntPhys2, linear probes recover more signal, and VideoMAE attains the strongest temporal attentive result at 73.9\% violation-of-expectation accuracy. On both benchmarks, attentive probe performance on LTX consistently underperforms compared to the other two pretraining paradigms.}
Together, these results suggest that intuitive-physics is broadly decodable from video foundation models, but it is not represented as a single, uniform capability. Its accessibility, and especially its temporal grounding, depends much more on the benchmark and readout than on a simple hierarchy of pretraining objectives or model scale.
\end{abstract}

\section{Introduction}
Video foundation models trained on internet scale data exhibit strong performance across tasks ranging from action recognition to future prediction and text-conditioned generation \citep{bardes2024revisitingfeaturepredictionlearning, videomaev22023, assran2025vjepa2selfsupervisedvideo, ltxvideo2024}. Yet, strong benchmark performance does not establish what kind of internal knowledge these models acquire, since apparent competence may be driven by superficial visual cues and benchmark-specific shortcuts rather than by representations that reflect physical understanding \citep{krojer2025shortcut, feng2025breakingvideollmbenchmarks}. \textcolor{black}{This motivates the central question of this paper: \textit{do pretrained video foundation models encode intuitive-physics information in their frozen representations, and if so, where in the network does it reside and how accessible is it?}} Many recent video architectures are implicitly or explicitly motivated as world models, systems that capture how objects and events evolve over time in a structured latent space \citep{bardes2024revisitingfeaturepredictionlearning, assran2023selfsupervisedlearningimagesjointembedding}. If such framing is accurate, we should expect pretrained representations to encode intuitive physical structure. \citet{garrido2025intuitivephysicsunderstandingemerges} offer evidence that this may be the case, showing that some aspects of intuitive physics already emerge in pretrained representations. At the same time, practical use cases often depend on temporal coherence \citep{ouyang2024codefcontentdeformationfields}, as well as on physical plausibility and robustness beyond benchmark-specific pattern matching. Yet as \citet{bordes2026intphys} and \citet{krojer2025shortcut} demonstrate, evaluating intuitive physics in modern video models remains difficult, because success on video benchmarks may reflect generic motion sensitivity, semantic recognition, or dataset biases rather than genuine sensitivity to physical constraints.

In this work, following the probing methodology of \citet{alain2018understandingintermediatelayersusing}, the frozen-feature analysis paradigm of \citet{El_Banani_2024_CVPR} and the more recent work of \citet{joseph2026interpretingphysicsvideoworld}, we use frozen-feature probing to ask not whether video foundation models can be adapted to solve intuitive-physics benchmarks, but what physics-relevant information is accessible in their frozen representations. Since probing measures decodability rather than whether a model actively uses that information, we compare readouts of increasing expressivity, using temporal controls to test the robustness of the recovered signal, and matched-backbone comparisons along with backbone sweeps to assess scale-related confounds.

This framing organizes our analysis around three research questions. \textcolor{black}{First, by comparing probes of increasing expressivity, linear, MLP, and temporal attentive, we assess the degree to which intuitive physics information is explicitly decodable from frozen features. Second, by probing multiple depths of each architecture, and multiple denoising stages for diffusion models, we ask where in the representational hierarchy this information becomes most accessible. Third, across three major pretraining paradigms: predictive joint-embedding learning, masked video reconstruction, and diffusion-based video generation, instantiated by V-JEPA \citep{bardes2024revisitingfeaturepredictionlearning, assran2025vjepa2selfsupervisedvideo, vjepa212026}, VideoMAE \citep{videomae2022, videomaev22023}, and LTX-Video \citep{ltxvideo2024}; we ask how the accessibility of intuitive-physics information differs under a common evaluation protocol and we further assess how much these differences can be explained by backbone scale through matched ViT-L comparisons, within-family backbone sweeps, and a comparison of LTX-Video 2B and 13B.}

As datasets to measure the decodability of intuitive physics we use two recent benchmarks that target physical understanding in video. IntPhys2 by \citet{bordes2026intphys} tests sensitivity to core principles such as permanence, immutability, spatio-temporal continuity, and solidity in controlled environments. MVP by \citet{krojer2025shortcut} complements this setting with minimal video pairs designed to reduce shortcut-based performance inflation: paired examples are visually similar but require opposite answers, making it harder to succeed through superficial cues alone. \textcolor{black}{Together, the two benchmarks capture complementary aspects of intuitive physics: IntPhys2 tests whether models detect violations of basic physical principles in controlled synthetic scenes, while MVP tests plausibility judgments on more natural videos, using a paired design that discourages shortcut solutions. Neither benchmark requires predicting exact dynamics or reasoning about counterfactuals, so our claims concern sensitivity to physical plausibility rather than physical reasoning in general.}

Code, experiment configurations, and saved results are available at
\url{https://github.com/Probe4Physics/Probe4Physics}.

\section{Related Work}
Related recent work has begun to study intuitive physics in pretrained video models from complementary perspectives. \citet{garrido2025intuitivephysicsunderstandingemerges} show that intuitive physics understanding can emerge from self-supervised pretraining on natural videos, particularly in models that predict in representation space, using violation-of-expectation style evaluation. \citet{joseph2026interpretingphysicsvideoworld} further examine how physical information is organized within V-JEPA 2 and VideoMAE-v2 through layerwise probing and interpretability analyses.

\subsection{Pretraining Paradigms in Video Foundation Models}
Self-supervised video representation learning has developed along several distinct pretraining objectives. One prominent line of work is masked video reconstruction, where models are trained to recover masked spatio-temporal content from partial observations. Architectures such as VideoMAE \citep{videomae2022} and its successor VideoMAE-v2 \citep{videomaev22023} show that reconstruction based objectives can learn strong and transferable video representations, even under very high masking ratios.

A second line of work replaces pixel reconstruction with latent space prediction. As shown by \citet{assran2023selfsupervisedlearningimagesjointembedding}, in the JEPA framework the model predicts target representations from context rather than reconstructing raw inputs, with the goal of encouraging more abstract and semantically structured representations. This idea was first developed by \citet{assran2023selfsupervisedlearningimagesjointembedding} in the visual domain with I-JEPA and later extended to video by \citet{bardes2024revisitingfeaturepredictionlearning} in V-JEPA, where prediction in representation space is explicitly connected to temporal understanding and, more broadly, to world modeling objectives. Recent variants such as V-JEPA 2 \citep{assran2025vjepa2selfsupervisedvideo} and V-JEPA 2.1 \citep{vjepa212026} further emphasize large scale learning and dense temporally grounded features.

A third family is diffusion-based video modeling~\citep{ltxvideo2024}, in which useful representations may arise implicitly through the denoising process used for video generation. Unlike masked reconstruction or latent prediction, diffusion models are not primarily optimized for representation quality, yet recent work suggests that their backbones can encode substantial motion information and coherent temporal-semantic structure \citep{xiao2024videodiffusionmodelstrainingfree, zhu2024exploringpretrainedtexttovideodiffusion}. In our experiments, we use LTX-Video as a representative of this family: a large-scale model designed for efficient real-time text-to-video generation, which makes it an informative counterpart to the representation-focused objectives of V-JEPA and VideoMAE.

Taken together, these paradigms optimize for different notions of structure: reconstruction, latent prediction, and denoising. However, it remains unclear how these differences affect the encoding and accessibility of intuitive physics information in pretrained video representations.

\subsection{Probing and Representation Evaluation}
Probing \citep{alain2018understandingintermediatelayersusing, probing} has become a standard tool for analyzing what information is accessible in learned representations. Early work on linear classifier probes introduced the idea of attaching independent readouts to intermediate layers in order to measure how task relevant information can be extracted across network depth. In this view, strong probe performance does not simply reflect end task accuracy, but offers a way to study how information is organized inside a pretrained model. In video representation learning, frozen feature evaluation has been widely used to assess whether pretrained backbones encode motion, temporal correspondence, and higher level semantic structure without full task specific finetuning \citep{bardes2024revisitingfeaturepredictionlearning, joseph2026interpretingphysicsvideoworld}. At the same time, the conclusions that can be drawn from probing depend strongly on probe expressivity. Linear probes test whether information is directly accessible, whereas stronger probes may recover information that is present but not explicitly organized for a linear readout. For this reason, comparing probe families of increasing capacity can be more informative than relying on a single probe type. Despite the broad use of probing for representation analysis, only a small number of recent studies have begun to examine intuitive-physics representations in pretrained video models, and these do not compare the main video pretraining paradigms under a unified layer-wise frozen-feature protocol \citep{garrido2025intuitivephysicsunderstandingemerges,joseph2026interpretingphysicsvideoworld}.

\subsection{Evaluating Intuitive Physics and Video Understanding}
Evaluating physical understanding in video is challenging because many benchmarks can be partially solved through superficial appearance cues, generic motion cues, or dataset specific biases. \citet{bordes2026intphys} created IntPhys2 to address this problem with controlled videos built around four core physical principles, namely permanence, immutability, spatio-temporal continuity, and solidity. Its possible versus impossible setup is designed to test sensitivity to physical plausibility rather than generic action recognition. MVP by \citet{krojer2025shortcut} addresses a complementary issue: shortcut-based score inflation in video question answering. It does so by constructing minimal video pairs that are visually similar and paired with the same question but opposite answers. Under this design, a model must answer both members of the pair consistently, making it much harder to rely on superficial visual or textual shortcuts.

These benchmarks provide a more targeted evaluation of intuitive physics than standard video understanding datasets alone. However, benchmark performance in isolation still does not fully separate sensitivity to physical structure from generic temporal modeling. This motivates controlled comparisons that examine not only overall accuracy, but also the conditions under which physics relevant information becomes accessible in pretrained representations.

\section{Models}
\textcolor{black}{
We compare three model families representing predictive joint-embedding learning (V-JEPA), masked video reconstruction (VideoMAE), and diffusion-based video generation (LTX-Video). Our main comparison uses the largest publicly available checkpoint of each model variant to characterize the strongest accessible representations from each family, rather than to isolate the causal effect of the pretraining objective. To assess scale-related confounds, we additionally compare matched ViT-L V-JEPA and VideoMAE models, perform within-family backbone sweeps, and compare LTX-Video 2B and 13B. These analyses test whether the observed differences persist when backbone size is matched or varied within a family. Training data and optimization remain unmatched, so we interpret the results as family-level associations rather than effects of the objective alone.
Detailed model descriptions are provided in Appendix \ref{app:models} and checkpoint/backbone configurations are summarized in \autoref{tab:models-overview}.}

\section{Datasets}

\subsection{IntPhys2}

IntPhys2 ~\citep{bordes2026intphys} organizes its videos around four physical principles (permanence, immutability, spatio-temporal continuity, and solidity) and structures them into scenes. Each scene consists of four clips: two that are physically possible and two that are impossible. This quadruplet structure is the unit of evaluation throughout our analysis \textcolor{black}{(see Appendix \autoref{fig:intphys2-conditions}).} The primary metric is Violation of Expectation (VOE) accuracy. A scene is counted as correct only when every possible clip in that scene receives a higher plausibility score than every impossible clip, i.e., $\min(\text{score}_\text{possible}) > \max(\text{score}_\text{impossible})$. This is stricter than per-clip accuracy: a model can classify the majority of clips correctly and still fail on VOE if it does not rank them consistently within a scene. 
Full filtering, split-count, stratification and adaptation are reported in Appendix ~\ref{app:dataset-details}.

\subsection{Minimal Video Pairs (MVP)}
\citet{krojer2025shortcut} proposed MVP, a shortcut-aware video question answering benchmark for physical understanding. Its key feature is the use of minimal video pairs: two videos are visually similar and associated with the same question, but require opposite answers. This structure is useful for our setting because it reduces the extent to which a model can succeed by  exploiting static or dataset-specific shortcuts. The main metric employed by the benchmark is pair consistency: a pair is counted as correct only when both members of the minimal pair are classified correctly. 

The original MVP task is formulated as video question answering, where each example is answered in the benchmark's local choice space. This creates a mismatch with our frozen-feature probing protocol. Most of the backbones considered in this work, such as VideoMAE and V-JEPA, are video-only representation models and do not natively accept a natural-language query. Although some video encoders can be aligned with a language model for video question answering, doing so would introduce an additional vision-language training stage and would no longer measure only the information already present in the frozen video representations. For this reason, we do not evaluate MVP as a full text-conditioned QA task, but instead adapt it to a binary physical-plausibility task. Appendix \autoref{fig:mvp-adaptation} summarizes this adaptation: the original text-conditioned VideoQA format is converted into a binary plausibility prediction task, and probe outputs are mapped back to the original answer space for pair-consistency scoring.

Concretely, we filter MVP to retain intuitive-physics examples whose questions can be mapped to a binary plausibility judgment, and require that each retained example has exactly two answer choices that can be semantically mapped to a yes-or-no option. Rather than training probes on the raw positional choice labels, which are not stable across examples, we convert each retained sample into a semantic binary label: plausible/possible versus implausible/impossible. At evaluation time, binary probe predictions are mapped back into the original MVP choice space so that scoring remains compatible with the benchmark's paired evaluation protocol. Full filtering, split-count, and stratification details are reported in Appendix~\ref{app:dataset-details}.

\section{Evaluation}

\textcolor{black}{To assess how intuitive-physics information is distributed across pretrained representations, we train probes on features extracted from multiple depths of each model. For every transformer-based video backbone, we select four relative depths corresponding approximately to 25\%, 50\%, 75\%, and 100\% of the network, so that models with different absolute numbers of layers can be compared under the same protocol. For example, a 24-layer ViT-L model is probed at layers 6, 12, 18, and 24, whereas a 40-layer ViT-G model is probed at layers 10, 20, 30, and 40. For LTX-Video, we apply the same four-depth principle over transformer blocks and additionally probe multiple denoising noise levels (10 steps linearly separated from 0.1 to 1.0), allowing us to analyze how intuitive-physics information changes across both network depth and the diffusion trajectory.} 

Because the linear and MLP probes require a single input vector, we perform average pooling over the spatio-temporal token embeddings produced by the backbones. We use average pooling rather than a \texttt{CLS} token for two reasons: first, not all evaluated architectures provide a directly comparable \texttt{CLS} representation; second, prior works of \citet{Pan_2021_ICCV, beyer2022betterplainvitbaselines} suggest that pooled token features can provide a stronger and more comparable summary for downstream evaluation.

Training and hyperparameter-selection details for all probe families are provided in Appendix~\ref{app:training}. Table~\ref{tab:probe_optuna} reports the Optuna search space for the linear and MLP probes, together with the learning-rate grid and fixed architectural settings used for the temporal attentive probe. Every configuration is trained with three random seeds (42, 101, and 102). Unless otherwise stated, we report the mean across these runs, with standard deviation indicating variability across them. To assess whether probe performance reflects physics-relevant information rather than static or visual cues, we evaluate temporally perturbed inputs for all probe families. We additionally train the temporal attentive probe with randomized labels to determine whether high performance can arise without a meaningful relationship between the representations and target labels. Full protocols and results are provided in Appendix Sections~\ref{app:temporal-controls}--\ref{app:random-control}.

\subsection{Probes}

\paragraph{Linear probe}
Following the standard probing setup introduced by \citet{alain2018understandingintermediatelayersusing}, these probes test whether task-relevant information is directly accessible from the extracted representation. Each linear probe consists of a single layer that maps the pooled clip embedding to the output space of the target task. As such, strong performance with this probe provides evidence that the relevant signal is encoded in a relatively explicit and linearly decodable form \citep{probing}.

\paragraph{Multi-layer perceptron probe}
To evaluate whether physics-relevant information is present but not linearly encoded, we also train multi-layer perceptron (MLP) probes. These extend the linear probe by introducing additional hidden layers interleaved with non-linear activations (GeLU) and normalization layers (LayerNorm). We vary the number of hidden layers, hidden dimensionality, and dropout rate, selecting the best configuration on the validation set. MLP probes provide a stronger readout than linear probes and can recover task-relevant signals that are not directly linearly decodable, although the interpretability of such results depends more strongly on probe capacity \citep{liu2019linguisticknowledgetransferabilitycontextual, hewitt2019designinginterpretingprobescontrol}.

\paragraph{Temporal attentive probe}
Finally, we evaluate a temporal attentive probe inspired by recent attentive probing protocols for transformer-based readouts over frozen video features \citep{bardes2024revisitingfeaturepredictionlearning, psomas2026attentionpleaserevisitingattentive, lin2022frozenclipmodelsefficient}. Unlike the linear and MLP probes, this probe operates directly on the sequence of token embeddings rather than on a pooled clip representation, allowing it to model temporal interactions explicitly. The probe consists of a shallow stack of self-attention layers followed by a final cross-attention layer used for classification. In our experiments, we use one self-attention layer and one final cross-attention layer, with 16 attention heads throughout. This probe therefore tests whether intuitive-physics information becomes more accessible when the readout mechanism can exploit the spatio-temporal structure directly.

\begin{table*}[t]
\centering
\small
\setlength{\tabcolsep}{3pt}
\renewcommand{\arraystretch}{1.08}
\begingroup
\newcommand{\runmetric}[2]{#1\,{\scriptsize$\pm$\,#2}}
\def\datasetsep{\smash{\rule[-0.6em]{0.4pt}{1.8em}}}
\resizebox{\textwidth}{!}{%
\begin{tabular}{lcc cc cc @{\hspace{0.65em}}c@{\hspace{0.65em}} cc cc cc}
\toprule
& \multicolumn{6}{c}{\textbf{IntPhys2}}
& \datasetsep
& \multicolumn{6}{c}{\textbf{MVP}} \\
\cmidrule(lr){2-7}
\cmidrule(lr){9-14}
\textbf{Model}
& \multicolumn{2}{c}{\textbf{Linear}}
& \multicolumn{2}{c}{\textbf{MLP}}
& \multicolumn{2}{c}{\textbf{Temporal Attn.}}
& \datasetsep
& \multicolumn{2}{c}{\textbf{Linear}}
& \multicolumn{2}{c}{\textbf{MLP}}
& \multicolumn{2}{c}{\textbf{Temporal Attn.}} \\
\cmidrule(lr){2-3}
\cmidrule(lr){4-5}
\cmidrule(lr){6-7}
\cmidrule(lr){9-10}
\cmidrule(lr){11-12}
\cmidrule(lr){13-14}
& \textbf{Layer} & \cellcolor{teal!18}\textbf{VOE}
& \textbf{Layer} & \cellcolor{teal!18}\textbf{VOE}
& \textbf{Layer} & \cellcolor{teal!18}\textbf{VOE}
& \datasetsep
& \textbf{Layer} & \cellcolor{teal!18}\textbf{Pair}
& \textbf{Layer} & \cellcolor{teal!18}\textbf{Pair}
& \textbf{Layer} & \cellcolor{teal!18}\textbf{Pair} \\
\midrule
V-JEPA
& 32 & \metriccol{\runmetric{48.4}{1.1}}
& 24 & \metriccol{\runmetric{45.8}{1.1}}
& 16 & \bestmetric{\runmetric{64.7}{5.2}}
& \datasetsep
& 24 & \metriccol{\runmetric{47.5}{0.4}}
& 32 & \metriccol{\runmetric{87.7}{1.0}}
& 24 & \bestmetric{\runmetric{95.0}{0.9}} \\

V-JEPA 2
& 40 & \metriccol{\runmetric{38.6}{2.3}}
& 30 & \metriccol{\runmetric{54.2}{3.0}}
& 40 & \bestmetric{\runmetric{66.0}{11.8}}
& \datasetsep
& 40 & \metriccol{\runmetric{47.9}{0.6}}
& 40 & \metriccol{\runmetric{86.3}{0.4}}
& 40 & \bestmetric{\runmetric{94.4}{0.4}} \\

V-JEPA 2.1
& 48 & \metriccol{\runmetric{36.6}{2.3}}
& 48 & \metriccol{\runmetric{39.9}{4.5}}
& 38 & \bestmetric{\runmetric{58.8}{5.2}}
& \datasetsep
& 38 & \metriccol{\runmetric{43.2}{1.3}}
& 38 & \metriccol{\runmetric{71.9}{1.3}}
& 48 & \bestmetric{\runmetric{92.5}{0.4}} \\

VideoMAE
& 24 & \metriccol{\runmetric{31.4}{3.4}}
& 24 & \metriccol{\runmetric{35.3}{5.9}}
& 24 & \bestmetric{\runmetric{73.9}{4.5}}
& \datasetsep
& 16 & \metriccol{\runmetric{35.8}{0.7}}
& 24 & \metriccol{\runmetric{64.7}{0.7}}
& 24 & \bestmetric{\runmetric{91.9}{0.2}} \\

VideoMAE-v2
& 40 & \metriccol{\runmetric{51.0}{3.9}}
& 20 & \metriccol{\runmetric{47.7}{3.0}}
& 20 & \bestmetric{\runmetric{58.8}{6.8}}
& \datasetsep
& 20 & \metriccol{\runmetric{36.0}{1.5}}
& 20 & \metriccol{\runmetric{66.4}{0.4}}
& 30 & \bestmetric{\runmetric{89.9}{3.3}} \\

LTX-Video
& \makecell{24 {\scriptsize (0.1)}} & \bestmetric{\runmetric{49.7}{3.0}}
& \makecell{24 {\scriptsize (0.2)}} & \metriccol{\runmetric{45.1}{3.4}}
& 24 {\scriptsize (0.2)} & \metriccol{\runmetric{48.4}{6.9}}
& \datasetsep
& \makecell{24 {\scriptsize (0.8)}} & \metriccol{\runmetric{50.2}{1.0}}
& \makecell{24 {\scriptsize (0.6)}} & \metriccol{\runmetric{68.6}{0.7}}
& \makecell{24 {\scriptsize (0.4)}} & \bestmetric{\runmetric{83.0}{2.9}} \\
\bottomrule
\end{tabular}%
}
\endgroup
\vspace{0.3em}
\caption{
Best-case benchmark comparison across largest model backbones. For each dataset
and probe, we report the best layer and mean test performance, with standard
deviation. Shaded columns denote the primary dataset-specific metric. Bold marks
the best probe for each model within each dataset. For LTX-Video, parentheses
indicate the denoising noise level.
}
\label{tab:exp1-bestcase-panels}
\end{table*}

\section{Results and Discussion}
\subsection*{Analysis 1: Decodability of physical information}
\label{sec:probe-expressivity}

\begin{researchquestionbox}
\textbf{Research question.}
Is intuitive-physics information linearly decodable, or only recoverable with a stronger probe?\end{researchquestionbox}

To answer this question, we compare best-layer performance across probe families, using the largest available backbone for all the models and the 13B variant of LTX-Video. \autoref{tab:exp1-bestcase-panels} shows how on MVP, probe capacity matters substantially: linear probes remain weak across all model families, whereas MLP and especially temporal attentive probes recover much stronger signal. For example, in the V-JEPA family, pair consistency rises from roughly 43--48\% with linear probes to 72--88\% with MLP probes, and to above 93\% with temporal attentive probes. VideoMAE and LTX-Video follow the same pattern, indicating that MVP-relevant physics information is only weakly accessible to a linear readout and becomes much more recoverable when the probe can model nonlinear or explicit temporal interactions. IntPhys2 shows a weaker and less uniform dependence on probe expressivity. Linear probes already recover moderate signal in several models, reaching around 48--51\% VOE for V-JEPA, VideoMAE-v2, and LTX-Video. Temporal-aware probes still help in most cases, but the gains are smaller and less consistent compared to the linear than on MVP.

\begin{takeawaybox}
\textbf{Takeaway}.
Intuitive-physics information is not equally explicit across tasks. On MVP, it is largely not directly readable from frozen features and requires stronger nonlinear or temporal readouts to become accessible. On IntPhys2, a larger fraction of the signal is already available to linear probes, but more expressive probes still often improve performance.
\end{takeawaybox}

\begin{figure*}[h]
  \centering
  \includegraphics[width=1\textwidth]{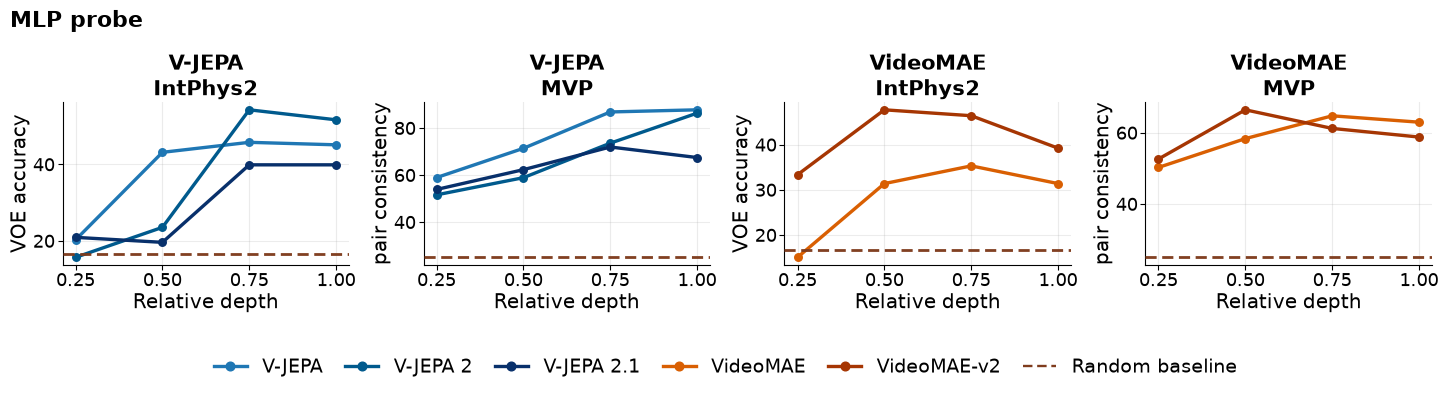}
    \caption{Layerwise MLP depth profiles by model family.}
  \label{fig:exp2_mlp}
\end{figure*}

\subsection*{Analysis 2: Layerwise emergence of physical information}
\label{sec:layerwise}

\begin{researchquestionbox}
\textbf{Research question.}
Where in the network does intuitive-physics information become most accessible?
\end{researchquestionbox}

To determine where physical understanding originates within the backbones, we analyzed the layer-wise performance profiles across models using their largest backbones available. Here, we focus our analysis on the MLP probe, as it possesses sufficient capacity to extract latent task-relevant signals without explicitly re-modeling temporal sequences. Corresponding layerwise profiles across all probe types for V-JEPA and VideoMAE are reported in Appendix \autoref{fig:exp2_all_probes}, while the LTX-Video linear layer--noise profile is shown in \autoref{fig:exp2_ltx_noise_linear}.

\textcolor{black}{As \autoref{fig:exp2_mlp} shows, the depth at which physics-relevant information becomes most accessible shows a similar trend between benchmarks and models. On both IntPhys2 and MVP performance generally improves toward middle-late layers (75\%), especially for the V-JEPA family: V-JEPA 2 rises from 51.56\% pair consistency on average at 0.25 depth to 86.31\% at the final layer on MVP (and goes from 16\% at 0.25 to 54.24\% average VOE at 0.75 on IntPhys2). VideoMAE models show the same broad mid-to-late trend, though with weaker gains and earlier peaks. LTX follows a similar pattern: as \autoref{fig:exp2_ltx_noise_mlp} in the Appendix illustrates its strongest physics signal is concentrated around the middle part of the backbone, and reducing noise does not produce a clear monotonic improvement. Performance fluctuates across the denoising trajectory, suggesting that intuitive-physics information is not progressively revealed in a simple stage-by-stage manner. However, even for diffusion features, the most accessible physics signal appears at intermediate depth rather than at the extremes of the denoising process.}

\begin{takeawaybox}
\textbf{Takeaway}.
Physics-relevant information follows a similar trend across models: it is weakest in early layers and becomes most accessible at intermediate-to-late depth. LTX follows the same broad pattern, with its strongest signal concentrated in the middle of the backbone.
\end{takeawaybox}

\begin{figure}[t]
    \centering
    \includegraphics[width=1\linewidth]{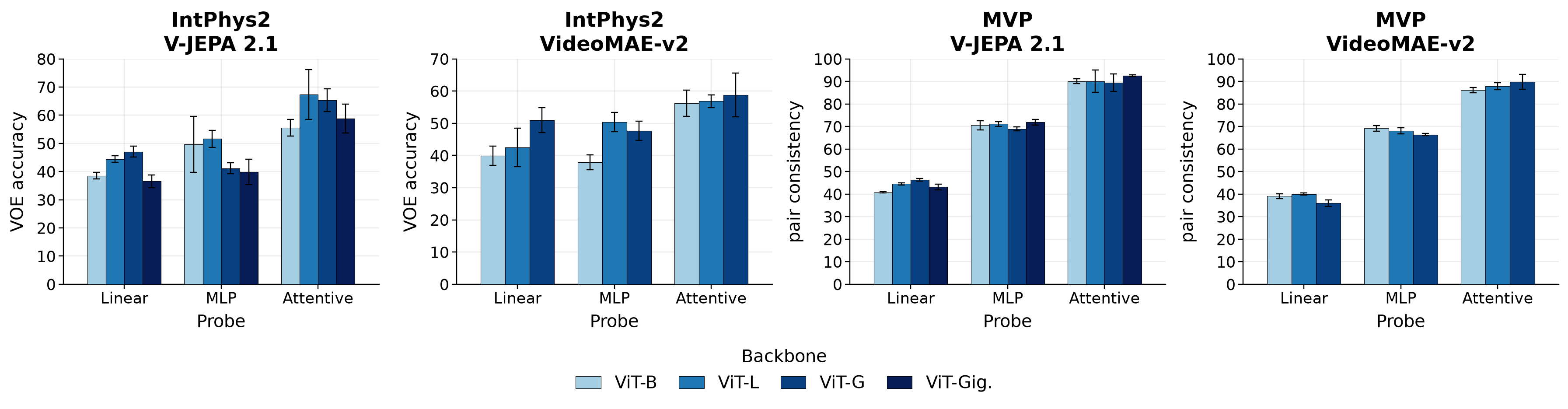}
    \caption{Backbone-scale comparison for V-JEPA 2.1 and VideoMAE-v2 across probes.}
    \label{fig:backbone_comparison_grouped}
\end{figure}

\begin{figure}[h]
    \centering
    \includegraphics[width=1\linewidth]{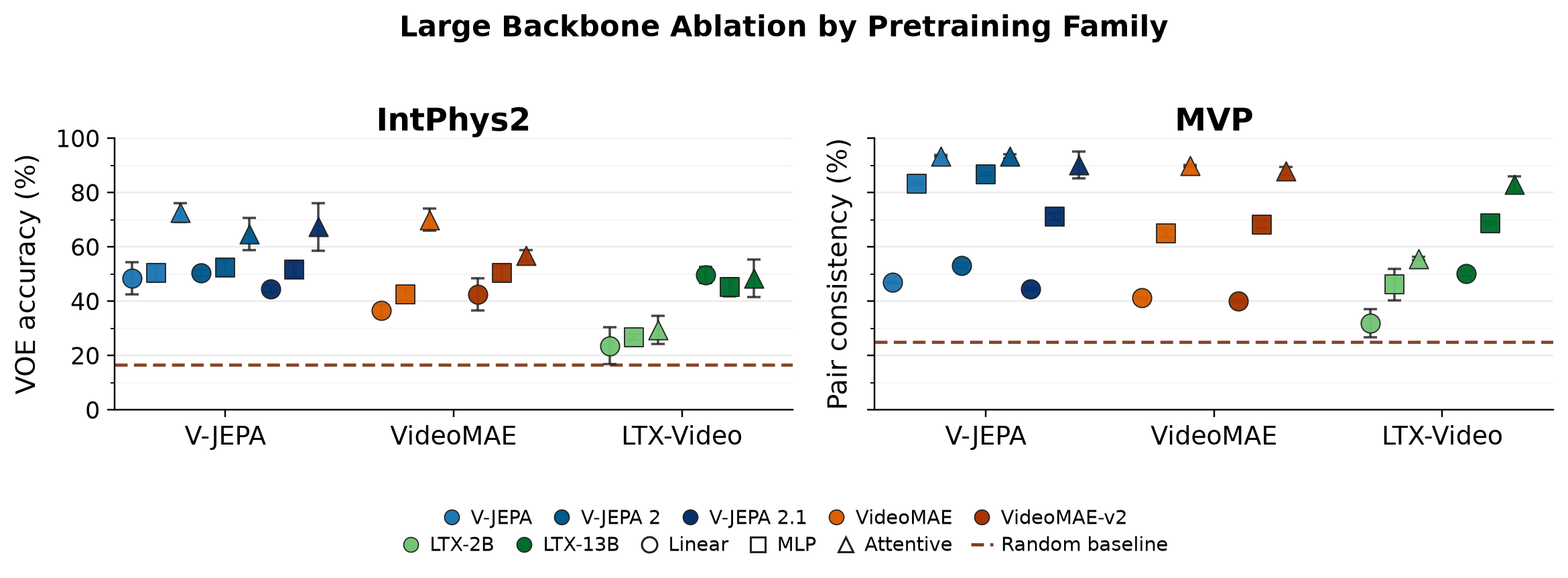}
    \caption{Best-case probe comparison across model families. V-JEPA and VideoMAE variants use matched ViT-L backbones, while LTX-Video is reported for both its 2B and 13B variants.}
    \label{fig:exp3_probe_comparison}
\end{figure}

\subsection*{Analysis 3: Best-case comparison across model families and backbones}
\label{sec:family-comparison}

\begin{researchquestionbox}
\textbf{Research question.}
How does accessibility differ across model families representing different pretraining paradigms?
\end{researchquestionbox}

To separate pretraining effects from backbone influence, we first compare the available backbone variants within V-JEPA 2.1, VideoMAE-v2, and LTX-Video. We support this comparison with pairwise Welch t-tests between backbone variants, using Holm correction within each dataset--model--probe group. \autoref{fig:backbone_comparison_grouped} shows that backbone scale can affect performance, especially for linear and MLP probes. However, this effect is not monotonic and depends strongly on both the benchmark and the probe type. 

On MVP, backbone size is generally not a decisive factor in the amount of physical information decoded by the probes. Most pairwise backbone comparisons are not significant after correction, although V-JEPA 2.1 with a linear probe shows significant differences between ViT-B and the larger ViT-L/ViT-G variants ($p_{\mathrm{Holm}} \approx 0.0058$ for both comparisons). For VideoMAE-v2, no MVP backbone comparison reaches statistical significance. On IntPhys2, the picture is less stable and harder to interpret: significant differences appear for the V-JEPA 2.1 linear probe ($p_{\mathrm{Holm}} = 0.0188$--$0.0400$) and the VideoMAE-v2 MLP probe ($p_{\mathrm{Holm}} =
  0.0171$--$0.0251$), but they do not support a clear "larger is better" trend.

Given this non-monotonic backbone effect, we then focus on a controlled best-case comparison across pretraining families. To reduce backbone-related confounds, we use the shared ViT-L backbone and compare families with the attentive probe, due to its ability to capture temporal interactions. \autoref{fig:exp3_probe_comparison} shows that there is no single dominant family. Both V-JEPA and VideoMAE encode physical information that is accessible to the attentive probe, and most pairwise comparisons between families are not statistically significant after correction. The main exceptions are the significant difference between V-JEPA and VideoMAE on MVP ($p_{\mathrm{Holm}} = 0.0091$), and the lower performance of LTX-2B, which differs significantly from several encoder-based models on both IntPhys2 ($p_{\mathrm{Holm}} = 0.0123$--$0.0295$) and MVP ($p_{\mathrm{Holm}} = 1.0\times10^{-5}$--$0.0025$).

\begin{takeawaybox}
    \textbf{Takeaway}. Intuitive-physics accessibility is not explained by backbone scale alone: scale effects are non-monotonic and probe-dependent. Instead, the results suggest that pretraining family and representation structure matter, with V-JEPA and VideoMAE generally exposing more accessible physics-relevant information than LTX in this setting.
\end{takeawaybox}

\section{Conclusions}

In this work, we studied how pretrained video foundation models encode intuitive-physics information and how this information is organized and recovered across model families, layers, backbone variants, and probes.

First, our probe-based analysis shows that accessibility depends strongly on the readout mechanism. On MVP, linear probes remain weak, whereas MLP and especially temporal attentive probes recover much stronger signal. On IntPhys2, useful information is more often accessible to linear probes, although expressive readouts still help. Second, our layerwise analysis shows that physics-relevant information is weakest in early layers and becomes most accessible at intermediate-to-late depth. LTX-Video follows a non-monotonic pattern, with its strongest signal appearing at particular intermediate blocks and noise levels. Third, our family and backbone comparisons do not identify a universally dominant model. V-JEPA achieves the strongest best-case result on MVP, while VideoMAE achieves the strongest temporal attentive result on IntPhys2. Increasing backbone size also does not produce a uniform improvement; its effect is model- and probe-dependent.

Taken together, our results show that pretrained video representations contain decodable physics-relevant information, but its accessibility depends on the benchmark, model family, representational depth, backbone variant, and readout mechanism.

\section{Limitations}
Our findings regard the accessibility of physics-relevant information in frozen representations rather than whether models actively use this information for physical reasoning. Comparisons across families remain influenced by differences in architecture, training data, and scale, while the higher capacity of the temporal attentive probe limits attribution of its gains to temporal modeling alone. Because we probe only four relative depths per backbone, the layerwise results identify broad representational regions rather than precise emergence points. Finally, our adapted MVP task and selected checkpoints and benchmarks cover only a subset of physical phenomena and video-modeling settings.

\bibliographystyle{plainnat}
\bibliography{references}

\clearpage
\appendix
\onecolumn

\section{Appendix}

\subsection{Code and Artifact Availability}

The implementation is publicly available at
\url{https://github.com/Probe4Physics/Probe4Physics}.
The repository includes data-preparation routines, frozen-feature
extraction, probe training and evaluation, experiment configurations,
saved measurements, and notebooks for generating the paper analyses.
Installation instructions, execution commands, and a table- and
figure-level reproduction index are provided in the repository.

Raw datasets, pretrained backbone weights, feature caches, and trained
probe checkpoints are not distributed with the code. Instructions for
obtaining the datasets and backbone weights, together with known
artifact-availability limitations, are documented in the repository.

\subsection{Dataset Construction and Additional Benchmark Details}\label{app:dataset-details}

\paragraph{IntPhys2}
We use the Main split of the dataset and discard the Debug split, which
contains simplified videos intended for model calibration and is not
representative of the evaluation conditions. This gives 253 scenes and
1{,}012 clips in total. We split at the scene level so that all four clips
of a quadruplet always fall in the same partition. The split is 60/20/20
and is stratified by physics condition to keep each principle proportionally
represented across train, validation, and test. The final counts are 151
scenes (604 clips) for training, and 51 scenes (204 clips) each for
validation and test. Each clip carries a native binary label (possible or
impossible) that maps directly onto the probe's binary output, so no label
conversion is required.

\begin{figure*}[h]
    \centering
    \includegraphics[width=0.98\textwidth]{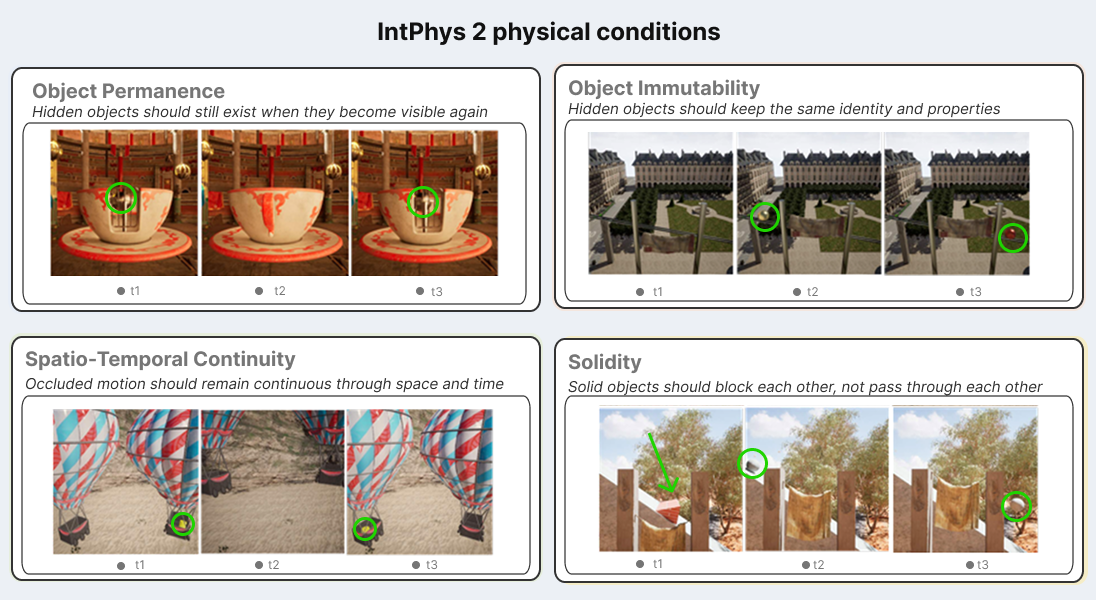}
    \caption{
    IntPhys2 physical conditions.
    Each panel shows an illustrative fixed-camera frame sequence for one principle:
    permanence, immutability, spatio-temporal continuity, and solidity.
    Frames are adapted from Fig.~6 of \citet{bordes2026intphys}; figure layout and annotations are ours.
    }
    \label{fig:intphys2-conditions}
\end{figure*}

\paragraph{MVP}

\begin{figure*}[t]
    \centering
    \includegraphics[width=0.95\textwidth]{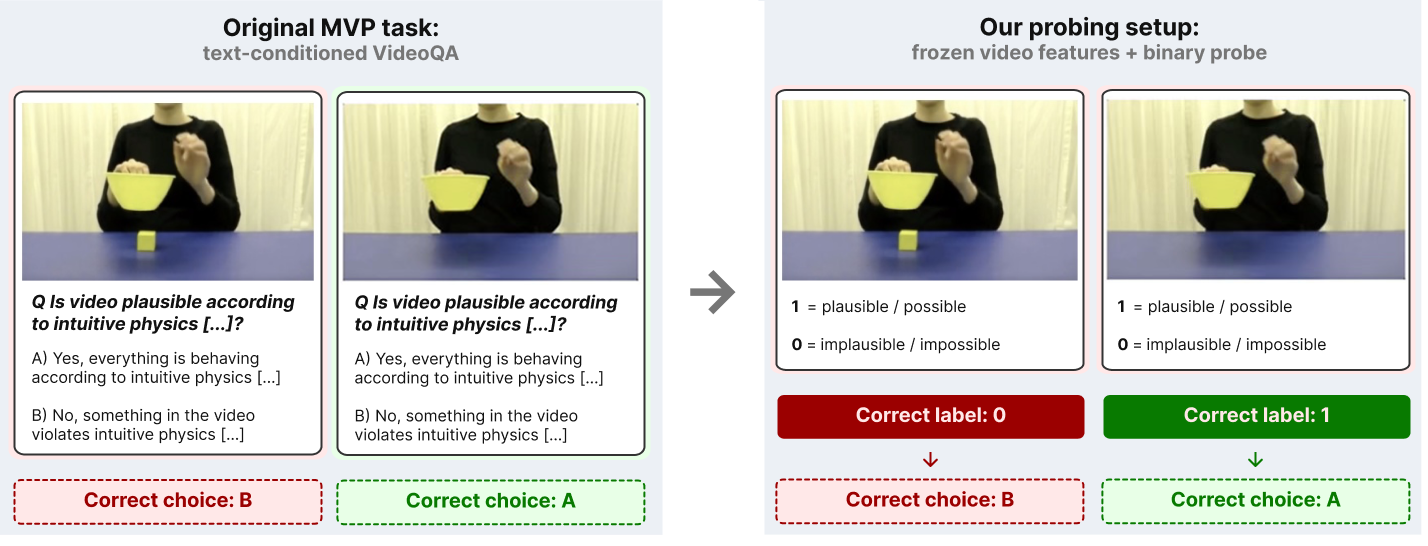}
    \caption{
    MVP adaptation for frozen-feature probing.
    We convert text-conditioned VideoQA examples from Minimal Video Pairs~\citep{krojer2025shortcut}
    into binary physical-plausibility labels, while preserving pair-consistency evaluation.
    Video frames are adapted from MVP examples; figure layout and annotations are ours.
    }
    \label{fig:mvp-adaptation}
\end{figure*}

We adapt MVP into a video-only binary plausibility task using a deterministic
selection rule. We first retain only examples from the intuitive-physics subset.
We then normalize each question by lowercasing, removing punctuation and other
non-alphanumeric characters, and collapsing whitespace. A question is retained
only if its normalized form contains one of a fixed set of plausibility-query
phrases, listed in \autoref{tab:mvp-template-groups}. These phrases cover
questions asking whether the video, event outcome, object trajectory, final
position, object location, or object interaction is physically plausible or
possible. We exclude questions whose answer space is not a binary plausibility
judgment, including counting questions and multi-choice future-prediction
questions; \autoref{tab:mvp-selection-summary} reports the resulting selection
counts.

After question filtering, we require each retained example to have exactly two
answer choices with yes/no-like semantics. We map answers corresponding to
yes, plausible, or possible to label 1, and answers corresponding to no,
implausible, or impossible to label 0. This mapping is semantic rather than
positional: both yes/no and no/yes choice orders occur in MVP and are handled
explicitly. The natural-language question is not provided to the probe; it is
used only for example selection and label construction. At evaluation time,
probe predictions are scored using the original MVP minimal-pair structure,
where a pair is counted as correct only if both videos in the pair are assigned
the correct binary label.

This procedure yields 4{,}943 minimal pairs, corresponding to 9{,}886 videos,
with exactly 4{,}943 plausible and 4{,}943 implausible examples. We split the
adapted dataset at the minimal-pair level using a 60/20/20
train/validation/test split, stratified by source and question-template group.
The resulting splits contain 2{,}965 training pairs, 989 validation pairs, and
989 test pairs. To reduce leakage risk, we verified that no minimal pair, exact
video reference, or conservative source-native event identifier appears in more
than one split. We release the retained pair identifiers, split assignments,
source labels, question-template groups, and sample identifiers with the
supplementary material.

\begin{table}[t]
\centering
\small
\setlength{\tabcolsep}{4pt}
\renewcommand{\arraystretch}{1.05}
\begin{tabular}{llr}
\toprule
\textbf{Source} & \textbf{Retained question type} & \textbf{Pairs} \\
\midrule
InfLevel & Video physically plausible/possible & 2{,}729 \\
GRASP & Final position of the ball plausible & 762 \\
GRASP & Trajectory of the ball plausible & 512 \\
GRASP & Outcome of the experiment plausible & 256 \\
GRASP & Trajectory of the rotating plank plausible & 128 \\
GRASP & Location of the cube plausible & 128 \\
GRASP & Final position of the top cube plausible & 128 \\
GRASP & Interaction between the balls plausible & 128 \\
IntPhys & Physical plausibility with rising wall context & 113 \\
IntPhys & General physical plausibility & 59 \\
\midrule
Total & & 4{,}943 \\
\bottomrule
\end{tabular}
\caption{
Retained MVP question-template groups for the adapted video-only plausibility
task. Counts are reported in minimal pairs.
}
\label{tab:mvp-template-groups}
\end{table}

\begin{table}[t]
\centering
\small
\setlength{\tabcolsep}{5pt}
\renewcommand{\arraystretch}{1.05}
\begin{tabular}{lr}
\toprule
\textbf{Selection outcome} & \textbf{Rows} \\
\midrule
Original MVP annotations & 54{,}828 \\
Outside intuitive-physics subset & 42{,}608 \\
Intuitive physics but not binary plausibility & 2{,}334 \\
Retained adapted examples & 9{,}886 \\
\midrule
Retained plausible/possible labels & 4{,}943 \\
Retained implausible/impossible labels & 4{,}943 \\
\bottomrule
\end{tabular}
\caption{
Selection summary for the adapted MVP task. Excluded intuitive-physics examples
include non-binary counting and prediction questions, such as questions asking
how many collisions occur or what event happens next.
}
\label{tab:mvp-selection-summary}
\end{table}

\paragraph{Comparability to original benchmarks.}
Our experiments adapt both datasets to a supervised frozen-feature probing
protocol. For IntPhys2, we construct train/validation/test splits at the scene
level in order to train and select probes. For MVP, we additionally convert the
original text-conditioned multiple-choice VideoQA task into a video-only binary
plausibility task. Therefore, the reported scores should be interpreted under
our probing protocol rather than as direct comparisons to published benchmark
scores.

\subsection{Extended Model Descriptions}
\label{app:models}
\paragraph{V-JEPA Family}
The V-JEPA (Video Joint-Embedding Predictive Architecture) \cite{bardes2024revisitingfeaturepredictionlearning} models leverage the principle of latent space prediction. Unlike reconstruction-based approaches that operate on pixel-level data, the model learns to predict a target representation from context in a shared embedding space. This encourages abstract, semantically meaningful, and compact representations rather than appearance matching. Compared with the original V-JEPA \cite{bardes2024revisitingfeaturepredictionlearning}, V-JEPA 2 \cite{assran2025vjepa2selfsupervisedvideo} and V-JEPA 2.1 \cite{vjepa212026} extend the same core idea to larger-scale training and stronger temporally grounded representations, with a greater emphasis on dense video understanding and world-model-like behavior. In our experiments, we use the largest available checkpoint within this family so that the probing results reflect the strongest pretrained representation we can access under the same evaluation protocol. 

\paragraph{VideoMAE Family}
The VideoMAE \cite{videomae2022} architecture represents the masked video reconstruction paradigm, where a model is trained to reconstruct masked spatio-temporal content from partial observations. Architecturally, it adapts the masked auto-encoding idea to video by using a ViT-based encoder on visible tokens and learning from a high masking ratio, which forces the model to exploit temporal context and long-range structure rather than local pixel continuity alone. VideoMAE-v2 \cite{videomaev22023} improves on the original formulation by strengthening representation learning and transfer performance, while keeping the same overall masked-reconstruction paradigm. Relative to v1, v2 is generally better at extracting transferable video features, so it provides a stronger baseline for probing whether reconstruction-based pretraining encodes intuitive physics. 

\paragraph{LTX Video Diffuser}
LTX Video belongs to the diffusion-based video generation family \cite{ltxvideo2024}. In this setting, the model learns to denoise a video sample over multiple steps, starting from a noisy latent and progressively refining it into a coherent video. The architecture is designed for generation rather than representation learning, but the denoising backbone can still encode useful information about motion, scene layout, and object interactions. Unlike V-JEPA and VideoMAE, whose representations are extracted from a single forward pass, LTX provides an additional dimension of analysis through the denoising trajectory itself, so we probe different diffusion stages to see when physical and temporal structure becomes most accessible. We evaluate the \texttt{ltxv-2b-0.9.8-distilled} and \texttt{ltxv-13b-0.9.8-distilled} checkpoints.

\begin{table*}[h]
  \centering
  \small
  \setlength{\tabcolsep}{4pt}

  \resizebox{\textwidth}{!}{%
  \begin{tabular}{@{} l c c !{\vrule width 0.8pt} l c c c r l @{}}
    \toprule
    \textbf{Family}
    & \textbf{Patch}
    & \textbf{Tubelet}
    & \textbf{Backbone/Variant}
    & \textbf{Res.}
    & \textbf{Depth}
    & \textbf{Probed layers}
    & \textbf{Params}
    & \textbf{Role} \\
    \midrule

    \multirow{2}{*}{V-JEPA}
      & \multirow{2}{*}{16}
      & \multirow{2}{*}{2}
      & ViT-L & 224 & 24 & 6, 12, 18, 24 & 305.49M & Matched encoder \\
      & & & ViT-H & 384 & 32 & 8, 16, 24, 32 & 637.55M & Main \\

    \specialrule{0.35pt}{0pt}{0pt}

    \multirow{2}{*}{V-JEPA 2}
      & \multirow{2}{*}{16}
      & \multirow{2}{*}{2}
      & ViT-L & 256 & 24 & 6, 12, 18, 24 & 303.89M & Matched encoder \\
      & & & ViT-g & 384 & 40 & 10, 20, 30, 40 & 1.01B & Main \\

    \specialrule{0.35pt}{0pt}{0pt}

    \multirow{4}{*}{V-JEPA 2.1}
      & \multirow{4}{*}{16}
      & \multirow{4}{*}{2}
      & ViT-B & 384 & 12 & 3, 6, 9, 12 & 86.83M & Scale sweep \\
      & & & ViT-L & 384 & 24 & 6, 12, 18, 24 & 304.68M & Matched encoder, Scale sweep \\
      & & & ViT-g & 384 & 40 & 10, 20, 30, 40 & 1.01B & Scale sweep \\
      & & & ViT-G & 384 & 48 & 12, 24, 38, 48 & 1.85B & Main, Scale sweep \\

    \specialrule{1.0pt}{0pt}{0pt}

    \multirow{2}{*}{VideoMAE}
      & \multirow{2}{*}{16}
      & \multirow{2}{*}{2}
      & ViT-L & 224 & 24 & 6, 12, 18, 24 & 303.90M & Matched encoder \\
      & & & ViT-H & 224 & 32 & 8, 16, 24, 32 & 631.61M & Main \\

    \specialrule{0.35pt}{0pt}{0pt}

    \multirow{3}{*}{VideoMAE-v2}
      & \multirow{3}{*}{16}
      & \multirow{3}{*}{2}
      & ViT-B & 224 & 12 & 3, 6, 9, 12 & 86.23M & Scale sweep \\
      & & & ViT-L & 224 & 24 & 6, 12, 18, 24 & 303.90M & Matched encoder, Scale sweep \\
      & & & ViT-g & 224 & 40 & 10, 20, 30, 40 & 1.01B & Main, Scale sweep \\

    \specialrule{1.0pt}{0pt}{0pt}

    \multirow{2}{*}{LTX-Video}
      & \multirow{2}{*}{1}
      & \multirow{2}{*}{1}
      & 2B & 224 & 28 & 7, 14, 21, 28 & 1.92B & Scale sweep \\
      & & & 13B & 224 & 48 & 12, 24, 36, 48 & 13.04B & Main \\

    \bottomrule
  \end{tabular}%
  }

 \caption{
    All backbone variants evaluated in this work (16 frames per clip throughout).
    Probed layers are reported as 1-based absolute block indices.
    For LTX-Video, each listed transformer block is probed across the
    10 configured noise levels.
    \textbf{Main}: largest checkpoint of each model variant, used in the main probe
    comparison and layerwise analysis.
    \textbf{Matched encoder}: ViT-L variants used for architecture-matched comparisons
    between the V-JEPA and VideoMAE families in Analysis~3.
    \textbf{Scale sweep}: within-family backbone variants used to assess the effect of
    model scale in Analysis~3.
    LTX-Video has no architecture-matched ViT counterpart; its 2B variant is included
    separately in the family comparison.
  }

  \label{tab:models-overview}
\end{table*}

\subsection{Model and Probe Configuration}\label{app:model-config}

\autoref{tab:models-overview} reports the model families, checkpoints/backbones, input resolutions, parameter counts, and probed depths used in each analysis. \autoref{tab:probe_optuna} lists the hyperparameter search space for the linear and MLP probes, together with the fixed settings used for the temporal attentive probe.

\begin{table*}[h]
\centering
\begin{tabular}{c c c} %
\toprule
Probe & Hyperparameter & Values \\

\midrule

Linear \& MLP & Learning Rate & Log-uniform in $[10^{-5}, 10^{-2}]$ \\
Linear \& MLP & Weight Decay & Log-uniform in $[10^{-8}, 10^{-2}]$ \\
Linear \& MLP & Batch Size & $\{32, 64, 128, 256\}$ \\
Linear \& MLP & Epochs & $\{20, 50, 100, 500, 1000, 2000\}$ \\

\midrule

MLP & Hidden dimensions & $\{[256], [512], [1024], [512,256],$ \\
& & $[1024,512,256], [1024,512,1024]\}$ \\
MLP & Dropout & Uniform in $[0, 0.5]$ \\

\midrule

Temporal Attentive & Learning Rates & $[5\mathrm{e}{-4}, 1\mathrm{e}{-4}, 5\mathrm{e}{-5}, 1\mathrm{e}{-5}]$ \\
Temporal Attentive &  Weight Decay & $0.01$ \\
Temporal Attentive &  Batch Size & $1$ \\
Temporal Attentive &  Epochs & $90$ (IntPhys2), $30$ (MVP) \\
Temporal Attentive &  Self Attention Layers & $1$ \\
Temporal Attentive &  Num Heads & $16$ \\
Temporal Attentive &  Dropout & $0.2$ \\
Temporal Attentive &  MLP Ratio & $2$ \\

\bottomrule
\end{tabular}

\vspace{0.5em}
\begin{minipage}{0.9\textwidth}
\centering

\end{minipage}

\caption{Hyperparameter optimization spaces for the linear and MLP probes,
and learning-rate grid and fixed settings for the temporal-attentive probe.}
\label{tab:probe_optuna}
\end{table*}

\subsection{Training Methodology}\label{app:training}

For all probe families, we keep the pretrained video backbone frozen and optimize only the probe parameters. We train a separate probe for each backbone--layer pair using the dataset splits produced by our benchmark pipeline (60\% train, 20\% validation, 20\% test). Optimization uses cross-entropy loss with AdamW \citep{loshchilov2019decoupledweightdecayregularization}. After every epoch, we evaluate on the validation split and retain the checkpoint with the highest validation benchmark-specific metric, breaking ties with lower validation loss. Hyperparameter selection is performed with a fixed random seed of 42. After selecting
the best configuration for each backbone--layer pair on the validation set, the selected
configuration is evaluated with three random seeds (42, 101, and 102) for the main
results. Temporal and randomized-label controls use seed 42 unless otherwise stated.

For the linear and MLP probes, we perform hyperparameter selection independently for each layer with Optuna \citep{optuna}. Each layer is assigned a 20-trial study with a TPE sampler \cite{TPE} and a median pruner. The sampler is seeded with 42, and pruning is enabled after 3 startup trials, with checks every epoch after a 100-epoch warm-up. The search space includes the learning rate, number of epochs, weight decay, batch size, and, for MLP probes, the hidden-layer configuration and dropout rate (\autoref{tab:probe_optuna}). Within each trial, we train on the training split, select the best checkpoint using the benchmark-specific downstream metric on the validation set. The best trial for each layer is then used for final evaluation on the held-out test split.

For the temporal attentive probe, we follow the same train/validation/test protocol as for the linear and MLP probes. Because token-level probing is substantially more memory-intensive than pooled-feature probing, we fix the batch size to 1 and tune only the learning rate via a small grid search over [$5\mathrm{e}{-4}, 1\mathrm{e}{-4}, 5\mathrm{e}{-5}, 1\mathrm{e}{-5}$], rather than running full Optuna-based optimization. For IntPhys2, all temporal attentive probes are trained for 90 epochs, using the same validation-based checkpoint selection criterion described above. The probe architecture is fixed to one self-attention block followed by a cross-attention block for classification, with 16 attention heads throughout.

\clearpage

\clearpage
\begin{figure}
    \centering
    \includegraphics[width=1\linewidth]{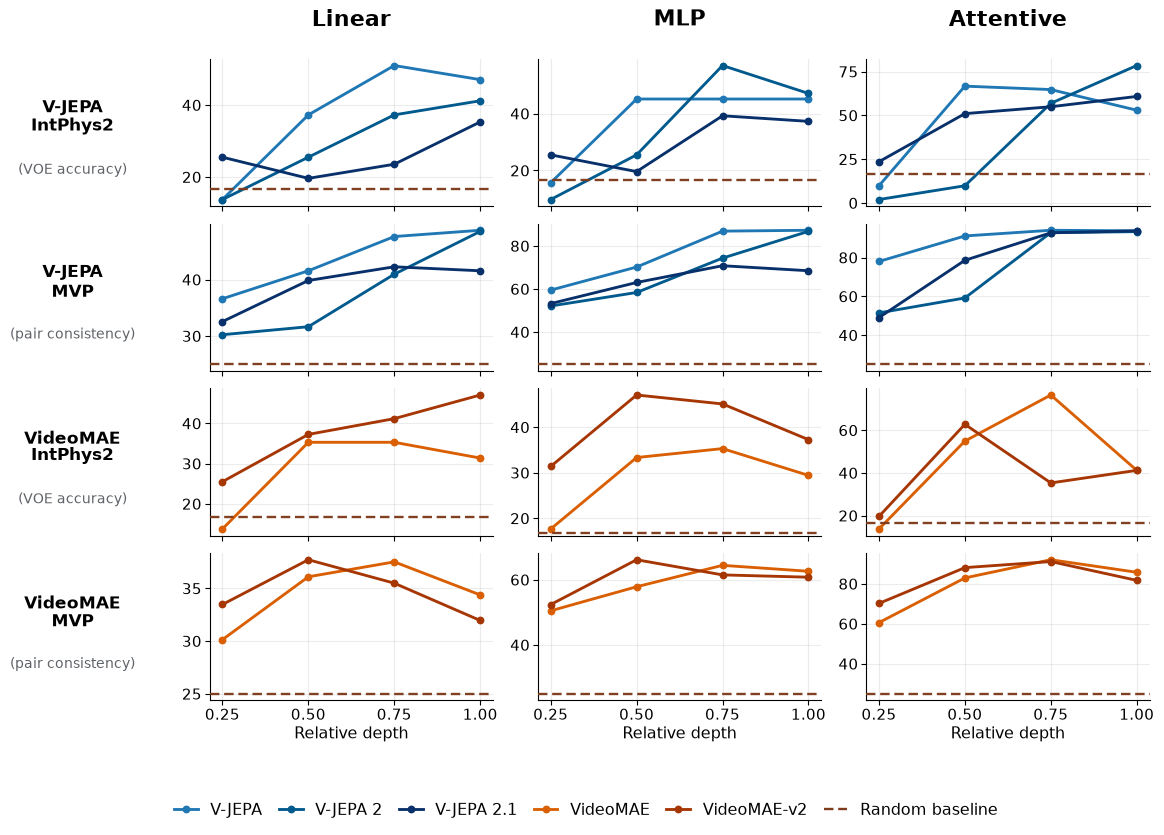}
    \caption{
Layerwise depth profiles for linear, MLP, and temporal attentive probes across
the V-JEPA and VideoMAE families. Dashed lines denote the random baselines.
}
    \label{fig:exp2_all_probes}
\end{figure}

\begin{figure}
    \centering
    \includegraphics[width=1\linewidth]{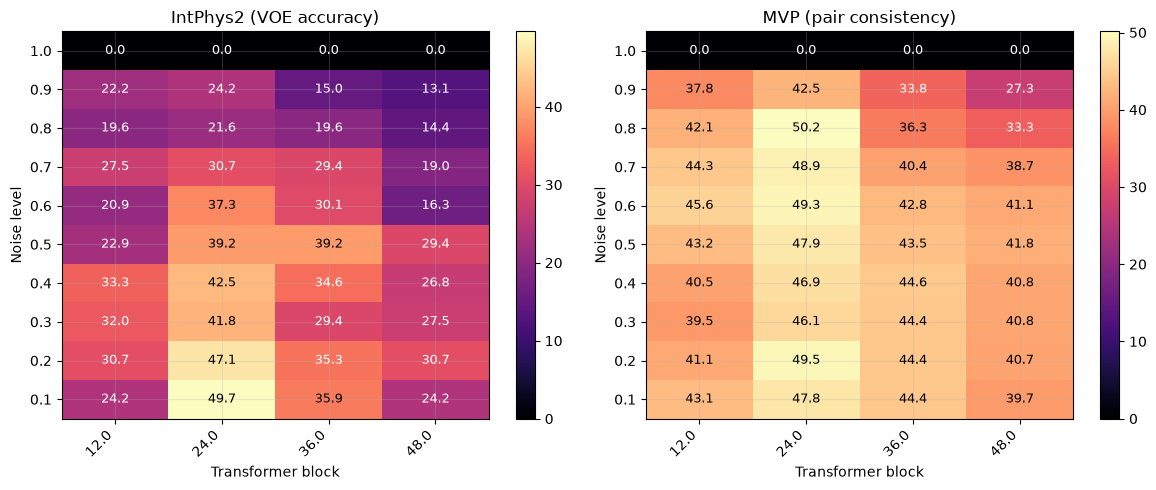}
    \caption{LTX-Video linear performance over denoising noise level and transformer block. The best LTX scores appear at specific denoising stages rather than uniformly across the trajectory}
    \label{fig:exp2_ltx_noise_linear}
\end{figure}

\begin{figure*}[t]
  \centering
  \includegraphics[width=\textwidth]{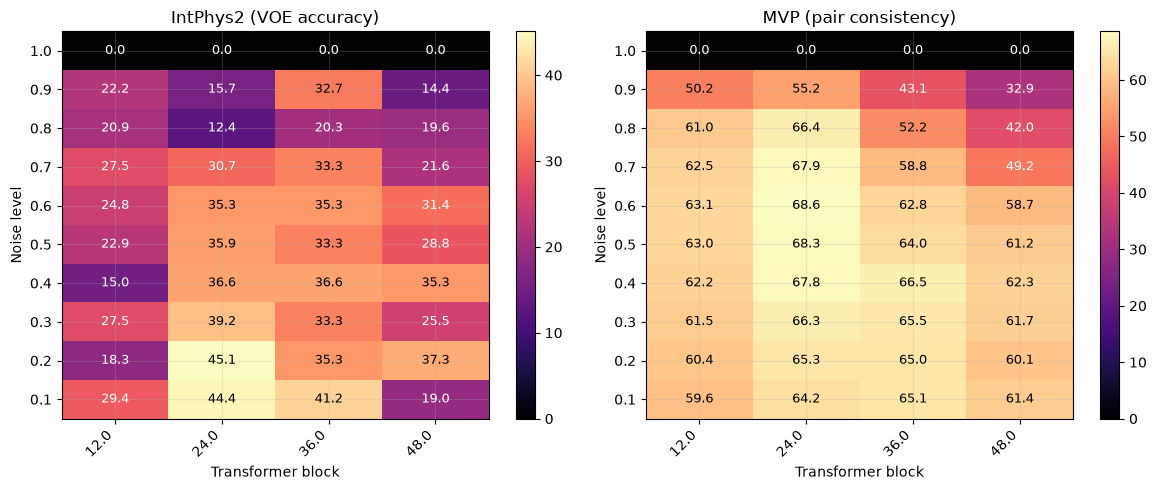}
  \caption{LTX-Video MLP performance over denoising noise level and transformer block. The best LTX scores appear at specific denoising stages rather than uniformly across the trajectory.}
  \label{fig:exp2_ltx_noise_mlp}
\end{figure*}

\begin{figure}
    \centering
    \includegraphics[width=1\linewidth]{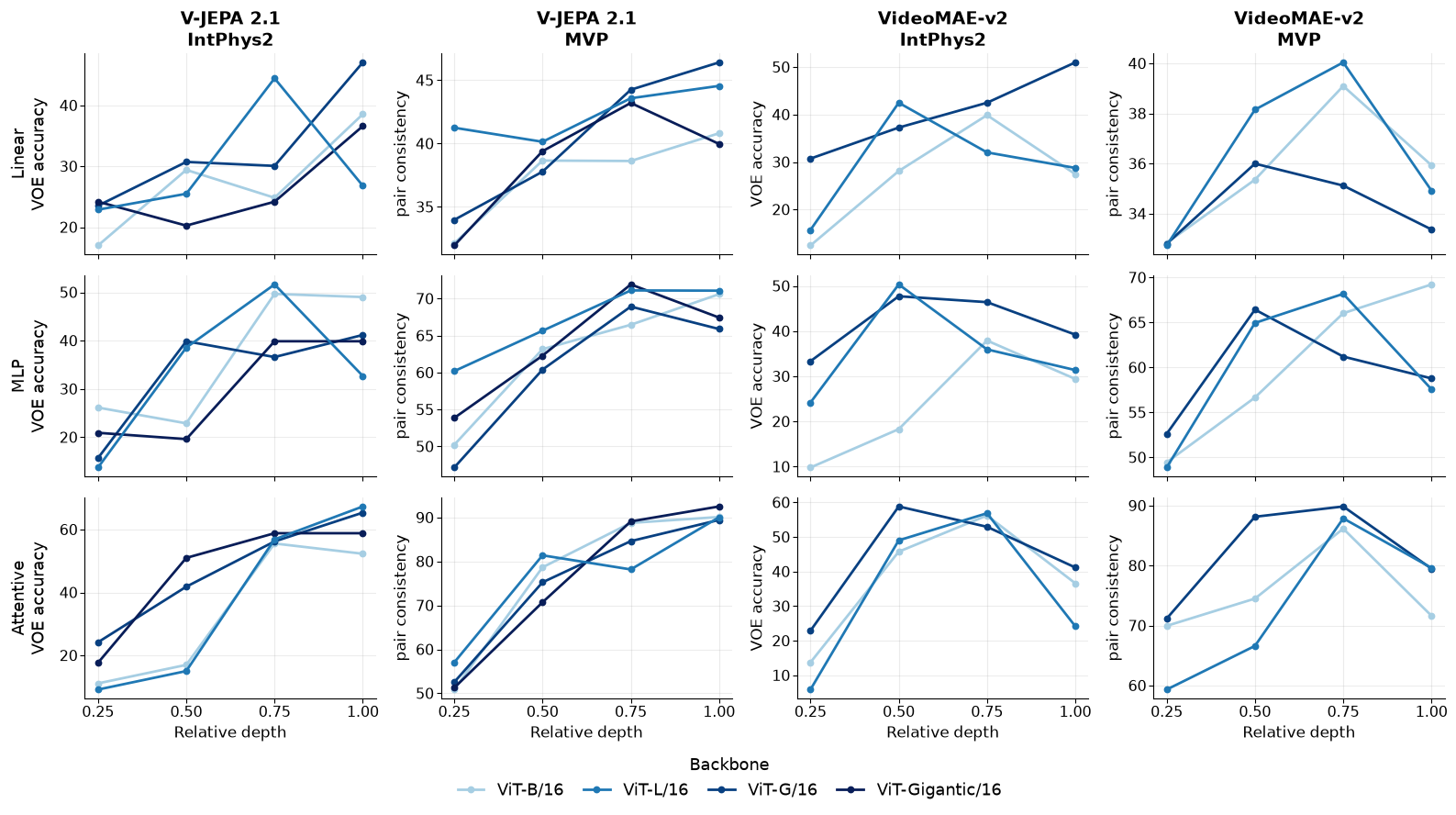}
    \caption{
Layerwise performance across backbone variants for V-JEPA 2.1 and VideoMAE-v2.
Rows correspond to linear, MLP, and temporal attentive probes, while columns
show results on IntPhys2 and MVP. Curves compare the available backbone sizes
over relative network depth.}
    \label{fig:placeholder}
\end{figure}

\clearpage
\subsection{Temporal Controls Analysis}    
\label{app:temporal-controls}
To verify that observed probe performance reflects genuine sensitivity to temporal physics dynamics rather than static appearance cues or order-invariant frame statistics, we evaluate each probe under two control conditions applied at the input level, keeping the probe and backbone frozen.

\paragraph{Frame-shuffled control} The temporal order of frames in each original clip is randomly permuted before feature extraction. This disrupts motion trajectories and causal event structure while preserving the set of frames and their individual appearance. A model whose performance is robust to shuffling cannot reliably be interpreted as relying on temporal ordering.

\paragraph{Single-frame control} 
A single frame is randomly sampled from each clip and repeated for the full clip length, producing a video of the same temporal dimension as the original input. This eliminates all temporal variation while keeping the input format identical to the main-task setting, thereby isolating static appearance as an upper bound on what can be inferred without any temporal dynamics.

For each model, probe type, and benchmark, we report the signed percentage-point change
in the primary metric with respect to the main-task score. Detailed results and discussion are provided in Appendix~\ref{app:temporal-controls}. These controls are essential for interpreting further results: if a model retains strong performance after temporal disruption, its main-task success may not reflect genuine sensitivity to physical dynamics.

\begin{figure}
    \centering
    \includegraphics[width=1\linewidth]{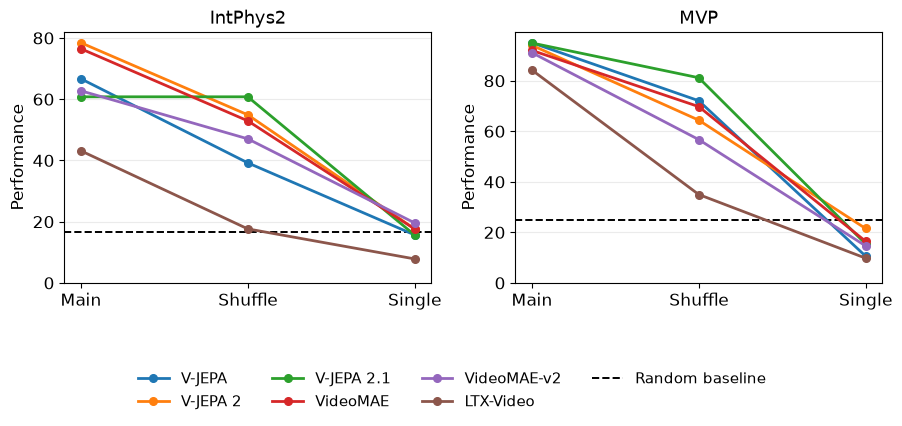}
    \caption{
Temporal attentive-probe performance under the main, frame-shuffled, and
single-frame conditions. Dashed lines indicate the random baselines.
}
\label{fig:temporal-controls-attentive}
\end{figure}

To assess whether probe performance depends on temporal structure rather than static appearance or unordered visual evidence, we evaluate the frame-shuffled
and single-frame controls introduced above.
\autoref{tab:temporal-controls} reports the signed percentage-point change for every
model and probe, while \autoref{fig:temporal-controls-attentive} shows the
absolute performance of the temporal attentive probes.

On MVP, temporal disruption causes substantial degradation. For MLP probes, frame
shuffling reduces pair consistency by 25.3--48.3 percentage points, while single-frame
repetition reduces it by 57.6--87.3 points. The same qualitative pattern holds for the
temporal attentive probes despite their high main-task scores. LTX-Video exhibits the
largest shuffle degradation under the temporal attentive probe ($-49.2$ points), while
V-JEPA exhibits the largest single-frame degradation ($-84.4$ points). MVP therefore
provides strong evidence that the recovered signal is not purely static: successful
minimal-pair classification depends substantially on multi-frame temporal evidence.

In contrast, IntPhys2 shows a more mixed reliance on temporal structure. Single-frame
repetition is consistently damaging across models and probe families, indicating that
the benchmark is not generally solvable from a single static frame. Frame shuffling,
however, is less uniformly destructive. In particular, both the linear and
temporal attentive probes for V-JEPA~2.1 show a $0.00$-point change under frame
shuffling. This suggests that, for these configurations, performance can be supported
by order-insensitive multi-frame information rather than the original temporal ordering.
This distinction is important for interpreting pooled probes: although linear and MLP
probes receive averaged features, the frozen backbone contextualizes tokens before
pooling and is therefore not necessarily invariant to temporal permutation.

Overall, temporal controls show that the probed signal is not purely static, especially
on MVP. However, IntPhys2 shuffle results suggest that some VOE performance can
come from order-insensitive plausibility cues rather than robust temporal
physical reasoning.

\begin{table}[t]
\centering
\textbf{(a) IntPhys2}\\[0.35em]
\resizebox{\textwidth}{!}{%
\begin{tabular}{lccc ccc ccc}
\toprule
\textbf{Model}
& \multicolumn{3}{c}{\textbf{Linear}}
& \multicolumn{3}{c}{\textbf{MLP}}
& \multicolumn{3}{c}{\textbf{Temporal Attn.}} \\
\cmidrule(lr){2-4}\cmidrule(lr){5-7}\cmidrule(lr){8-10}
& \textbf{Main} & \makecell{\textbf{$\Delta\!\downarrow$ shuffle}\\\textbf{(pp)}} & \makecell{\textbf{$\Delta\!\downarrow$ single}\\\textbf{(pp)}}
& \textbf{Main} & \makecell{\textbf{$\Delta\!\downarrow$ shuffle}\\\textbf{(pp)}} & \makecell{\textbf{$\Delta\!\downarrow$ single}\\\textbf{(pp)}}
& \textbf{Main} & \makecell{\textbf{$\Delta\!\downarrow$ shuffle}\\\textbf{(pp)}} & \makecell{\textbf{$\Delta\!\downarrow$ single}\\\textbf{(pp)}} \\
\midrule
V-JEPA      & 50.98 & \textbf{-29.41} & -43.14 & 45.10 & -25.49 & -27.45 & 66.67 & \textbf{-27.45} & -50.98 \\
V-JEPA 2    & 41.18 & -9.80 & -15.69 & 56.86 & \textbf{-33.33} & \textbf{-35.29} & 78.43 & -23.53 & \textbf{-60.78} \\
V-JEPA 2.1  & 35.29 & 0.00 & -25.49 & 39.22 & -9.80 & -19.61 & 60.78 & 0.00 & -45.10 \\
VideoMAE    & 35.29 & -15.69 & -25.49 & 35.29 & -21.57 & -21.57 & 76.47 & -23.53 & -58.82 \\
VideoMAE-v2 & 47.06 & -25.49 & -31.37 & 47.06 & \textbf{-33.33} & \textbf{-35.29} & 62.75 & -15.69 & -43.14 \\
LTX-Video   & 49.02 & \textbf{-29.41} & \textbf{-49.02} & 47.06 & -31.37 & -25.49 & 43.14 & -25.49 & -35.29 \\
\bottomrule
\end{tabular}
}
\vspace{1.0em}

\textbf{(b) MVP}\\[0.35em]
\resizebox{\textwidth}{!}{%
\begin{tabular}{lccc ccc ccc}
\toprule
\textbf{Model}
& \multicolumn{3}{c}{\textbf{Linear}}
& \multicolumn{3}{c}{\textbf{MLP}}
& \multicolumn{3}{c}{\textbf{Temporal Attn.}} \\
\cmidrule(lr){2-4}\cmidrule(lr){5-7}\cmidrule(lr){8-10}
& \textbf{Main} & \makecell{\textbf{$\Delta\!\downarrow$ shuffle}\\\textbf{(pp)}} & \makecell{\textbf{$\Delta\!\downarrow$ single}\\\textbf{(pp)}}
& \textbf{Main} & \makecell{\textbf{$\Delta\!\downarrow$ shuffle}\\\textbf{(pp)}} & \makecell{\textbf{$\Delta\!\downarrow$ single}\\\textbf{(pp)}}
& \textbf{Main} & \makecell{\textbf{$\Delta\!\downarrow$ shuffle}\\\textbf{(pp)}} & \makecell{\textbf{$\Delta\!\downarrow$ single}\\\textbf{(pp)}} \\
\midrule
V-JEPA      & 48.74 & -13.95 & -43.78 & 87.26 & \textbf{-48.33} & \textbf{-87.26} & 95.05 & -22.95 & \textbf{-84.43} \\
V-JEPA 2    & 48.53 & -17.59 & \textbf{-46.31} & 86.75 & -46.31 & -80.99 & 94.03 & -29.63 & -72.50 \\
V-JEPA 2.1  & 42.26 & -14.96 & -40.55 & 70.78 & -34.78 & -65.82 & 94.94 & -13.75 & -79.47 \\
VideoMAE    & 37.51 & -8.09 & -32.56 & 64.41 & -25.28 & -59.96 & 92.01 & -22.24 & -75.53 \\
VideoMAE-v2 & 37.71 & -9.61 & -34.88 & 66.13 & -29.22 & -63.90 & 91.10 & \textbf{-34.38} & -76.44 \\
LTX-Video   & 50.56 & \textbf{-25.99} & -41.25 & 68.86 & -39.53 & -57.63 & 84.23 & -49.24 & -74.42 \\
\bottomrule
\end{tabular}
}
\caption{
Temporal-control results across models, probes, and benchmarks. Main is the
seed-42 performance at the best main-condition layer, measured by VOE accuracy
on IntPhys2 or pair consistency on MVP. Control columns report the signed
percentage-point change at the same layer,
$\Delta_{\mathrm{pp}}=\mathrm{Control}-\mathrm{Main}$.
More negative values indicate stronger degradation under temporal disruption;
bold marks the largest degradation within each dataset, probe type, and control.
}
\label{tab:temporal-controls}
\vspace{0.4em}
\end{table}

\clearpage
\subsection{Random Labels Control}    
\label{app:random-control}

We use a random-label control to test whether the temporal attentive probe can
obtain high scores simply because it is expressive enough to fit the benchmark.
This control is applied only to the temporal attentive probe, since this is the
most powerful readout in our setup and therefore the one most exposed to this
criticism.

Concretely, we run the same training and evaluation pipeline leaving the backbone
features, data splits, selected layer, and probe architecture unchanged; only
the training labels are replaced. For MVP, labels are shuffled within each
minimal pair. For IntPhys2, labels are shuffled within each scene. This preserves
the benchmark structure and class balance inside each pair or scene, while
breaking the correspondence between the video representation and its true plausibility label.

If the attentive probes were merely learning dataset artifacts or memorizing the
evaluation structure, they should still perform well under this control. Instead,
as shown in \autoref{tab:random-label-control}, performance collapses toward
chance, indicating that the main attentive-probe results depend on the true
label structure rather than on probe capacity alone.

\begin{table*}[htbp]
\centering
\footnotesize
\setlength{\tabcolsep}{3pt}
\renewcommand{\arraystretch}{1.05}
\begingroup
\def\datasetsep{\smash{\rule[-0.6em]{0.4pt}{1.8em}}}
\resizebox{\textwidth}{!}{%
\begin{tabular}{l c ccc ccc @{\hspace{0.55em}}c@{\hspace{0.55em}} c ccc ccc}
\toprule
& \multicolumn{7}{c}{\textbf{IntPhys2}}
& \datasetsep
& \multicolumn{7}{c}{\textbf{MVP}} \\
\cmidrule(lr){2-8}
\cmidrule(lr){10-16}
\textbf{Model}
& \textbf{Layer}
& \multicolumn{3}{c}{\textbf{VOE}}
& \multicolumn{3}{c}{\textbf{Accuracy}}
& \datasetsep
& \textbf{Layer}
& \multicolumn{3}{c}{\textbf{Pair}}
& \multicolumn{3}{c}{\textbf{Accuracy}} \\
\cmidrule(lr){3-5}
\cmidrule(lr){6-8}
\cmidrule(lr){11-13}
\cmidrule(lr){14-16}
&

& \textbf{Main}
& \textbf{Rand.}
& $\boldsymbol{\Delta_{\mathrm{pp}}}$
& \textbf{Main}
& \textbf{Rand.}
& $\boldsymbol{\Delta_{\mathrm{pp}}}$
& \datasetsep
& \textbf{}
& \textbf{Main}
& \textbf{Rand.}
& $\boldsymbol{\Delta_{\mathrm{pp}}}$
& \textbf{Main}
& \textbf{Rand.}
& $\boldsymbol{\Delta_{\mathrm{pp}}}$ \\
\midrule
V-JEPA
& 16 & 66.67 & 13.73 & -52.94 & 78.92 & 51.96 & -26.96
& \datasetsep
& 24 & 95.05 & 10.11 & -84.93 & 97.27 & 50.61 & -46.66 \\

V-JEPA 2
& 40 & 78.43 & 11.76 & \textbf{-66.67} & 82.84 & 50.00 & \textbf{-32.84}
& \datasetsep
& 40 & 94.03 & 8.80 & \textbf{-85.24} & 96.51 & 49.24 & -47.27 \\

V-JEPA 2.1
& 48 & 60.78 & 19.61 & -41.18 & 72.55 & 44.12 & -28.43
& \datasetsep
& 38 & 94.94 & 16.99 & -77.96 & 97.12 & 49.04 & \textbf{-48.08} \\

VideoMAE
& 24 & 76.47 & 13.73 & -62.75 & 80.88 & 50.49 & -30.39
& \datasetsep
& 24 & 92.01 & 13.45 & -78.56 & 95.10 & 50.05 & -45.05 \\

VideoMAE-v2
& 20 & 62.75 & 7.84 & -54.90 & 72.55 & 50.00 & -22.55
& \datasetsep
& 30 & 91.10 & 14.66 & -76.44 & 94.89 & 50.61 & -44.29 \\

LTX-Video
& \makecell{48 {\scriptsize (0.2)}} & 43.14 & 9.80 & -33.33 & 64.22 & 48.53 & -15.69
& \datasetsep
& \makecell{24 {\scriptsize (0.4)}} & 84.23 & 12.03 & -72.19 & 91.25 & 49.44 & -41.81 \\
\bottomrule
\end{tabular}%
}
\endgroup
\caption{
Random-label control for the temporal attentive probe. Main is the seed-42
performance at the best main-condition layer. Rand. reports performance after
retraining the same probe with randomized labels at the same layer, and
$\Delta_{\mathrm{pp}}=\mathrm{Rand.}-\mathrm{Main}$.
}
\label{tab:random-label-control}
\vspace{0.4em}
\end{table*}

\subsection{Computational Requirements}
\label{app:compute}

Feature extraction was performed once for each dataset--backbone pair and the
resulting features were cached and reused across all probe types. For the largest
checkpoints considered, extraction required $0.38$--$0.83$ GPU-hours on IntPhys2
and $2.07$--$6.63$ GPU-hours on MVP, depending on the backbone
(\autoref{tab:feature-extraction-compute}).

Probe training used different hardware depending on the readout. Linear and MLP
probes were trained on CPU nodes, since they operate on pooled clip features and
are comparatively lightweight. Temporal attentive probes were trained on GPUs,
because they operate directly on token sequences. As a representative example,
\autoref{tab:probe-compute} reports the compute required by V-JEPA 2.1 with a
ViT-G backbone over the four probed layers used in the main experiments. In this
setting, the linear and MLP searches required $0.11$--$2.24$ CPU node-hours
across datasets, whereas the temporal attentive learning-rate search required
$23.4$--$85.7$ A100 GPU-hours. This confirms that the temporal attentive probe is
the dominant cost in the probing pipeline.

All encoder feature-extraction jobs and temporal attentive-probe jobs used
single A100 GPUs. LTX-Video feature extraction used single H100 GPUs. Linear and
MLP probe searches were run on single CPU nodes.

\begin{table}[t]
\centering
\small
\begin{tabular}{lllc}
\toprule
\textbf{Dataset} & \textbf{Model} & \textbf{Checkpoint} & \textbf{Extraction GPU-h} \\
\midrule
IntPhys2 & V-JEPA      & ViT-H/16 & 0.490 \\
IntPhys2 & V-JEPA 2    & ViT-g/16 & 0.608 \\
IntPhys2 & V-JEPA 2.1  & ViT-G/16 & 0.789 \\
IntPhys2 & VideoMAE    & ViT-H/16 & 0.379 \\
IntPhys2 & VideoMAE-v2 & ViT-g/16 & 0.454 \\
IntPhys2 & LTX-Video   & LTX-13B  & 0.831 \\
MVP      & V-JEPA      & ViT-H/16 & 3.471 \\
MVP      & V-JEPA 2    & ViT-g/16 & 4.105 \\
MVP      & V-JEPA 2.1  & ViT-G/16 & 5.729 \\
MVP      & VideoMAE    & ViT-H/16 & 2.070 \\
MVP      & VideoMAE-v2 & ViT-g/16 & 2.697 \\
MVP      & LTX-Video   & LTX-13B  & 6.633 \\
\bottomrule
\end{tabular}
\caption{Feature-extraction compute for the largest checkpoint of each model family. Features are extracted once per dataset--backbone pair and reused across probes.}
\label{tab:feature-extraction-compute}
\end{table}

\begin{table}[t]
\centering
\small
\begin{tabular}{llcccc}
\toprule
\textbf{Dataset} & \textbf{Probe}
& \textbf{GPU-h/layer}
& \textbf{GPU-h/4 layers}
& \textbf{CPU node-h/layer}
& \textbf{CPU node-h/4 layers} \\
\midrule
IntPhys2 & Linear    & 0.000 & 0.000  & 0.028 & 0.112 \\
IntPhys2 & MLP       & 0.000 & 0.000  & 0.101 & 0.405 \\
IntPhys2 & Attentive & 5.841 & 23.365 & 0.000 & 0.000 \\
MVP      & Linear    & 0.000 & 0.000  & 0.103 & 0.412 \\
MVP      & MLP       & 0.000 & 0.000  & 0.560 & 2.239 \\
MVP      & Attentive & 21.414 & 85.657 & 0.000 & 0.000 \\
\bottomrule
\end{tabular}
\caption{Probe training and search compute for a representative largest-backbone encoder setting: V-JEPA 2.1 ViT-G evaluated over four probed layers.}
\label{tab:probe-compute}
\end{table}

\clearpage

\end{document}